\documentclass{article} 

\usepackage{enumerate}
\usepackage{amsmath}
\usepackage{amsfonts}
\usepackage{amssymb}
\usepackage{amsbsy}
\usepackage{amsthm}
\usepackage{algorithm}
\usepackage{algorithmic}
\usepackage{mathrsfs}
\usepackage{paralist}
\usepackage{epsfig}
\usepackage{subfig}
\usepackage{makeidx} 

\usepackage[numbers,sort&compress]{natbib}
\usepackage{url}

\iftrue
\theoremstyle{plain} 
\theoremstyle{plain} 
\theoremstyle{plain} 
\theoremstyle{plain} 
\theoremstyle{plain} 
\theoremstyle{plain} \newtheorem{lemma}{Lemma}[section]
\theoremstyle{plain} 
\theoremstyle{plain} 
\theoremstyle{plain} 

\fi

\def\Naturals {{\mathbb{N}}}

\def\CM {{\mathcal{M}}}

\def\CX {{\mathcal{X}}}

\renewcommand \Pr {{\mbox{\bf{P}}}}

\newcommand \defn {\mathrel{\triangleq}}

\newcommand \ind{\mathop{\mbox{\ensuremath{\mathbb{I}}}}}

\begin{document}

\title{Statistical Decision Making for Authentication and Intrusion Detection}

\author{Christos Dimitrakakis\\
  Informatics Institute,\\ University of Amsterdam, \\ Science Park 107, NL-1098XG Amsterdam\\ c.dimitrakakis@uva.nl\\
\and
Aikaterini Mitrokotsa\\
Faculty of EEMCS\\
Delft University of Technology\\ Mekelweg 4, NL-2628CD Delft\\ 
A.Mitrokotsa@TUDelft.nl\\
}

\maketitle

\begin{abstract}
  User authentication and intrusion detection differ from standard
  classification problems in that while we have data generated from
  legitimate users, impostor or intrusion data is scarce or
  non-existent.  We review existing techniques for dealing with this
  problem and propose a novel alternative based on a principled
  statistical decision-making view point.  We examine the technique on
  a toy problem and validate it on complex real-world data from an
  RFID based access control system.  The results indicate that it can
  significantly outperform the classical world model approach.  The
  method could be more generally useful in other decision-making
  scenarios where there is a lack of adversary data.
  
\end{abstract}

\section{Introduction}

Classification is the problem of categorising data in one of two or
more possible classes.  In the classical {\em supervised learning}
framework, examples of each class have already been obtained and the
task of the decision maker is to accurately categorise new
observations, whose class is unknown. The accuracy is either measured
in terms of the rate of misclassification, or in terms of the average
cost, for problems where different types of errors carry different
costs.  In that setting, the problem has three phases:
\begin{inparaenum}[(a)]
\item the collection of training data,
\item the estimation of a decision rule based on the training data and
\item the application of the decision rule to new data.
\end{inparaenum}
Typically, the decision rule remains fixed after the second
step. Thus, the problem becomes that of finding the decision rule with
minimum risk from the training data.

Unfortunately, some problems are structured in such a way that it is
not possible to obtain data from all categories to form the decision
rule.  Novelty detection, user authentication, network intrusion
detection and spam filtering all belong to this type of decision
problems: while the data of the ``normal'' class is relatively easily
characterised, the data of the other class which we wish to detect is
not. This is partially due to the potentially adversarial nature of
the process that generates the data of the alternative class.  As an
example, consider being asked to decide whether a particular voice
sample belongs to a specific person, given a set of examples of his
voice and your overall experience concerning the voices of other
persons.  

In this paper, we shall employ two conceptual classes: the ``user''
and the ``adversary''.  The main distinction is that while we shall
always have examples of instances of the user class, we may not have
any data from the adversary class.

This problem is alleviated in authentication settings, where we must
separate accesses by a specific user from accesses by an adversary.
Such problems contain additional information: data which we have
obtained from other people.  This can be used to create a {\em world
  model}, which can then act as an adversary model, and has been used
with state-of-the-art results in
authentication~\cite{furui,cardinaux,reynolds:adaptedGMM}.

Since there is no explicit adversary model, the probability of an
attack cannot be estimated.  Our main contribution is a decision
making principle which employs a {\em pessimistic estimate} on the
probability of an attack.  Intuitively, this is done by conditioning
the adversary model on the {\em current} observations, whose class is
unknown. This enables us to place an upper bound on the probability of
the adversary class, in a \textit{Bayesian} framework. To the best of
our knowledge, this is the first time that such a Bayesian worst-case
approach has been described in the literature. The proposed method is
compared with both an \textit{oracle} and the \textit{world model}
approach on a test-bench. This shows that our approach can outperform
the world model under a variety of conditions. This result is
validated on the real-world problem of detecting unauthorised accesses
in a building.

The remainder of this section discusses related work. The model
framework is introduced in Sec.~\ref{sec:models}, with the proposed
Bayesian estimates discussed in Sec.~\ref{sec:bayesAdversary} and
methods for estimating the prior in Sec.~\ref{sec:world-model}. The
conclusion is preceded by Sec.~\ref{sec:experiments}, which presents
experiments and results.

\subsection{Related work}
\label{sec:related-work}

Classification algorithms have been extensively used for the detection
of intrusions in wired~\cite{mukkamala,cardenas} and
wireless~\cite{liu,deng} networks.  Their main disadvantage is that
labelled normal and attack data must be available for training.  After
the training phase, the classifier's learnt model will be used to
predict the labels of new unknown data.
However, such data is very hard to obtain and often unreliable. 
Finally, there will always exist new unknown attacks for which
training data are not available at all.

Outlier detection~\cite{ramaswamy,breunig} and
clustering~\cite{portnoy} use unlabelled data and are in
principle able to detect unknown types of attacks.  The main
disadvantage is that no explicit adversarial model is employed.

An alternative framework is the world model
approach~\cite{furui,cardinaux,reynolds:adaptedGMM}.  This is
extensively used in speech and image authentication problems, where
data from a considerable number of users are collected
to create a world model (also called a universal background model).
This approach is closely related to the model examined in this paper,
since it originates in the seminal work of~\cite{Gauvain94maximuma},
who employed an empirical Bayes technique for estimating a prior over
models.  Thus, the world model is a distribution over models, although
due to computational considerations a point estimate is used instead
in practice \cite{reynolds:adaptedGMM}.

The adversary may actively try to avoid detection, through knowledge
of the detection method. In essence, this changes the setting from a
statistical to an adversarial one.  For such problems, game theoretic
approaches are frequently used.  Dalvi et al. \cite{Dalvi}
investigated the adversarial classification problem as a two-person
game. More precisely, they examined the optimal strategy of an
adversary against a standard (adversary-unaware) classifier as well as
that of a classifier (adversary-aware) against a rational adversary.
This was under the assumption that the adversary has complete
knowledge of the detection algorithm.  In a similar vein, Lowd et
al. \cite{lowd} have investigated algorithms for reverse engineering
linear classifiers.  This allows them to retrieve sufficient
information to mount effective
attacks. 

In our paper we do not consider repeated interactions and thus we do
not follow a game-theoretic approach.  We instead consider how to
model the adversary, when we have a lot of data from legitimate users,
but no data from the adversary.  Our main contribution is a Bayesian
method for calculating a subjective upper bound on attack
probabilities without any knowledge of the adversary model. This can
be obtained simply by using the current (unlabelled) observations to
create a worst-case (or more generally pessimistic) model of the
adversary.\footnote{Some simpler alternative approaches are explored
  in an accompanying technical
  report~\cite{dimitrakakis:tr-uva-09-02}.} This is done by
conditioning the prior over adversary models according to new
(unlabelled) observations.

However, in order to control overfitting, we first condition the
adversary model's prior on the data of the remaining population of
users. This results in an empirical Bayes estimate of the
prior~\cite{Robbins:EmpiricalBayes:1955}, which is what the world
model approach essentially is~\cite{Gauvain94maximuma}. The prior then
acts as a soft constraint when selecting the worst-case adversary
model.

It is worthwhile to note that the problem of constructing a model for
a class with no data is related to the problem of null hypothesis
testing, for which similar ideas have appeared. For example,
\cite{edwards:bayesian-inference-psychological} explored the idea of
constructing a maximum likelihood estimate from the obsrvations and
using this as the alternative hypothesis. More sophisticated examples
for simple parametric problems were examined in
\cite{Berger:PointNull-P-Values}. This involved selecting the
worst-case prior from a given class of priors in order to be maximally
pessimistic about the null hypothesis. Our approach is similar in
spirit, but the application and technical details are substantially
different.

Our final contribution is an experimental analysis on a
synthetic problem, as well as on some real-world data, with promising
results: we show that the widely used {\em world model} approach
cannot outperform the proposed model.

\section{The proposed model framework}
\label{sec:models}

In the framework we consider, we assume that the set of all possible
models is $\CM$. Each model $\mu$ in $\CM$ is associated with a
probability measure over the set of observations $\CX$, which will be
denoted by $\mu(x)$ for $x \in \CX$, $\mu \in \CM$, so long as there
is no ambiguity. We must decide whether some observations $x \in \CX$,
have been generated by a model $q$ (the user) or a model $w$ (the
adversary) in $\CM$. Throughout the paper, we assume a prior
probability of the user having generated the data, $\Pr(q)$, with a
complementary prior $\Pr(w) = 1 - \Pr(q)$, for the adversary.

In the easiest scenario, we have perfect knowledge of $q, w \in
\CM$. It is then trivial to calculate the probability $\Pr(q|x)$ that
the user $q$ has generated the data $x$. This is the {\em oracle}
decision rule, defined in section~\ref{sec:oracle}. This is not a
realisable rule, as although we could accurately estimate $q$ with
enough data, in general there is no way to estimate the adversary
model $w$.

We thus consider the case where the user model is known and where we
are given a prior density $\xi(w)$ over the possible adversary models
$w \in \CM$. Currently seen observations are then used to form a
pessimistic posterior $\xi'$ for the adversary. This is explained in
Section~\ref{sec:bayesAdversary}.

Section~\ref{sec:world-model} discusses the more practical case where
neither the user model $q$, nor a prior $\xi$ over models $\CM$
are known, but must be estimated from data.  More precisely, the
section discusses methods for utilising other user data to obtain a
prior distribution over models.  This amounts to an empirical Bayes
estimate of the prior distribution~\cite{Robbins:EmpiricalBayes:1955}.
It is then possible to estimate $q$ by conditioning the prior on the
user data.  This is closely related to the adapted world model
approach~\cite{reynolds:adaptedGMM}, used in authentication
applications, which however, usually employs a point approximation to
the prior~\cite{cardinaux}.

\subsection{The oracle decision rule}
\label{sec:oracle}

We shall measure the performance of all the models against that of the
{\em oracle} decision rule.  The oracle enjoys perfect information
about the distribution of both the user and the adversary, and thus
knows both $q$ and $w$, as well as the {\em a priori} probability of
an attack, $\Pr(w)$.  On average, no other decision rule can do
better.

More precisely, let $\CM$ be the space of all models.  Let the adversary's
model be $w$ and the user's model be $q$, with $q,w \in \CM$.  Given
some data $x$, we would like to determine the probability that the
data $x$ has been generated by the user, $\Pr(q|x)$.  The oracle model
has knowledge of $w,q$ and $\Pr(q)$, so using Bayes' rule we obtain:
\begin{equation}
  \label{eq:oracle}
  \Pr(q|x) = \frac{q(x)\Pr(q)}{q(x) \Pr(q) + w(x)(1-\Pr(q))}.
\end{equation}
However, we usually have uncertainty about both the adversary and the
user model.  Concerning the adversary, the uncertainty is much more
pronounced.  The next section examines a model for the probability of
an attack when the user model is perfectly known but we only have a
prior $\xi(w)$ for the adversary model.

\subsection{Bayesian adversary model}
\label{sec:bayesAdversary}
We can use a subjective prior probability $\xi(w)$ over possible
adversary models, to calculate the probability of observations given
that they have been generated by the adversary: $\xi(x) = \int_{\CM}
w(x) \xi(w) dw$.\footnote{Here we used the fact that $\xi(x|w) =
  w(x)$, since the probability of the observations given a specific
  model $w$ no longer depends on our belief $\xi$ about which model
  $w$ is correct.}
Given a user model $q$, we can express the probability of the user $q$
given the observations $x$ under the belief $\xi$ as:
\begin{equation}
  \label{eq:bayesDecision}
  \xi(q | x) \defn \Pr(q |x, \xi)
  = \frac{q(x) \Pr(q)}{q(x) \Pr(q) +  \xi(x)\left(1-\Pr(q)\right)}.
\end{equation}
The difference with \eqref{eq:oracle} is that, instead of $w(x)$, we
use the marginal density $\xi(x)$. If $\xi(w)$ represents our
subjective belief about the adversary model $w$, then
\eqref{eq:bayesDecision} can be seen as the Bayesian equivalent of the
world model approach, where the prior over $w$ plays the role of the
world model.  Now let: $\xi'(w) \defn \xi(w|x)$ be the model posterior
for some observations $x$.  We shall need the following lemma:
\begin{lemma}
  For any probability measure $\xi$ on $\CM$, where $\CM$ is a space
  of probability distributions on $\CX$, such that each $\mu \in \CM$
  defines a probability (density) $\mu(x)$ with $x \in \CX$, with admissible
  posteriors $\xi'(\mu) \defn \xi(\mu | x)$, the marginal likelihood
  satisfies:
    $\xi'(x) \geq \xi(x)$, $\forall x \in \CX$.
    \label{lem:posterior-inequality}
\end{lemma}
A simple proof, using the Cauchy-Schwarz inequality on the norm
induced by the measure $\xi$, is presented in the Appendix.  From the
above lemma, it immediately follows that:
\begin{align}
  \label{eq:bayesBound}
  \xi(q|x)
  &\geq \xi'(q|x)
  =\frac{q(x) \Pr(q)}{q(x) \Pr(q) +  \left(1-\Pr(q)\right) \int_\CM w(x) \xi'(w) \, dw},
\end{align}
since $\xi'(x) = \int w(x) \xi'(w) \, dw \geq \int w(x) \xi'(w) \, dw
= \xi(x)$.  Thus \eqref{eq:bayesBound} gives us a subjective upper
bound on the probability of the data $x$ having been generated by the
adversary. This bound can then be used to make decisions.
Finally, note that we can form $\xi'(w)$ on a {\em subset} of
$x$. This possibility is explored in the experiments.

\subsection{Prior and user model estimation}
\label{sec:world-model}
Specifically for user authentication, we have data from two
sources. The first is data collected from the user which we wish to
identify. The second is data collected from other
persons.\footnote{These are not necessarily other users.} The $i$-th
person can be fully specified in terms of a model $\mu_i \in \CM$,
with $\mu_i$ drawn from some unknown distribution $\gamma$ over
$\CM$. If we had the models $\mu_i \in \CM$ for all the other people
in our dataset, then we could obtain an {\em empirical} estimate
$\hat{\gamma}$ of the prior distribution of models. Empirical Bayes
methods for prior estimation~\cite{Robbins:EmpiricalBayes:1955} extend
this procedure to the case where we only observe $x \sim \mu_i$, data drawn from the model $\mu_i$.

Let us now apply this prior over models to the estimation of the
posterior over models for some user. Given an estimate $\hat{\gamma}$
of $\gamma$, and some data $x \sim \mu$ from the user, and assuming
that $\mu \sim \gamma$, we can form a posterior for $\mu$ using Bayes
rule: $\hat{\gamma}(\mu|x) = \mu(x) \xi(\mu) / \int_{\CM}
\hat{\gamma}(x|\mu) \xi(d\mu)$, over all $\mu \in \CM$.  For a
specific user $k$ with data $x_k$, we write the posterior as
$\psi_k(\mu) \defn \hat{\gamma}(\mu|x_k)$.  Whenever we must decide
the class of a new observation $x$, we set the prior over the
adversary models to $\xi = \hat{\gamma}$ and then condition on part,
or all, of $x$ to obtain the posterior $\xi'(w)$. We then calculate
\begin{equation}
  \Pr(q_k | x,
  \xi', \psi_k)= \frac {\psi_k(x) \Pr(q_k)} {[\psi_k(x) \Pr(q_k) + (1-\Pr(q_k))
  \xi'(x)]}, 
\end{equation}
the posterior probability of the $k$-th user given the observations
$x$ and our beliefs $\xi'$ and $\psi_k$ over adversary and user {\em
  models} respectively.  When $\xi' = \xi$, we obtain an equivalent to
the world model approach of \cite{reynolds:adaptedGMM}, which is an
approximate form of the empirical Bayes procedure suggested
in~\cite{Gauvain94maximuma}.

\subsubsection{Prior estimation for multinomial models}
\label{sec:multinomial-estimation}
For discrete observations, we can consider multinomial distributions
drawn from a Dirichlet density, and use a maximum likelihood estimate
based on Polya distributions for $\gamma$.  More specifically, we use
the fixed point approach suggested in~\cite{minka2003edd} to estimate
Dirichlet parameters $\Phi$ from a set of multinomial observations.

To make this more concrete, consider multinomial observations of
degree $K$.  Our initial belief $\xi(\mu)$ is a Dirichlet prior with
parameters $\Phi \defn (\phi_1, \ldots, \phi_K)$ over models:
$\xi(\mu) = \frac{1}{B(\Phi)} \prod_{i=1}^K \mu_i^{\phi_i -1}$, which
is conjugate to the multinomial~\cite{Degroot}.  Given a sequence of
observations $x_1, \dotsc, x_n$, with $x_t \in 1, \ldots K$, where
each outcome $i$ has fixed probability $\mu_i$, then $c_i =
\sum_{t=1}^n \ind(x_t = i)$, where $\ind$ is an indicator function, is
multinomial and the posterior distribution over the parameters $\mu_i$
is also Dirichlet with parameters $\phi_i' = \phi_i + c_i$.  The
approach suggested in~\cite{minka2003edd} uses the following fixed
point iteration for the parameters: $\phi^{new}_i = \phi_i
\frac{\sum_k \Psi(c_{ik} + \phi_i) - \Psi(\phi_i)}{\sum_k \Psi(c_k +
  \sum_i \phi_i) - \Psi(\sum_i a_i)}$, where $\Psi(\cdot)$ is the
digamma function.
\begin{figure}[htb]
  \centering
\includegraphics[width=\textwidth]{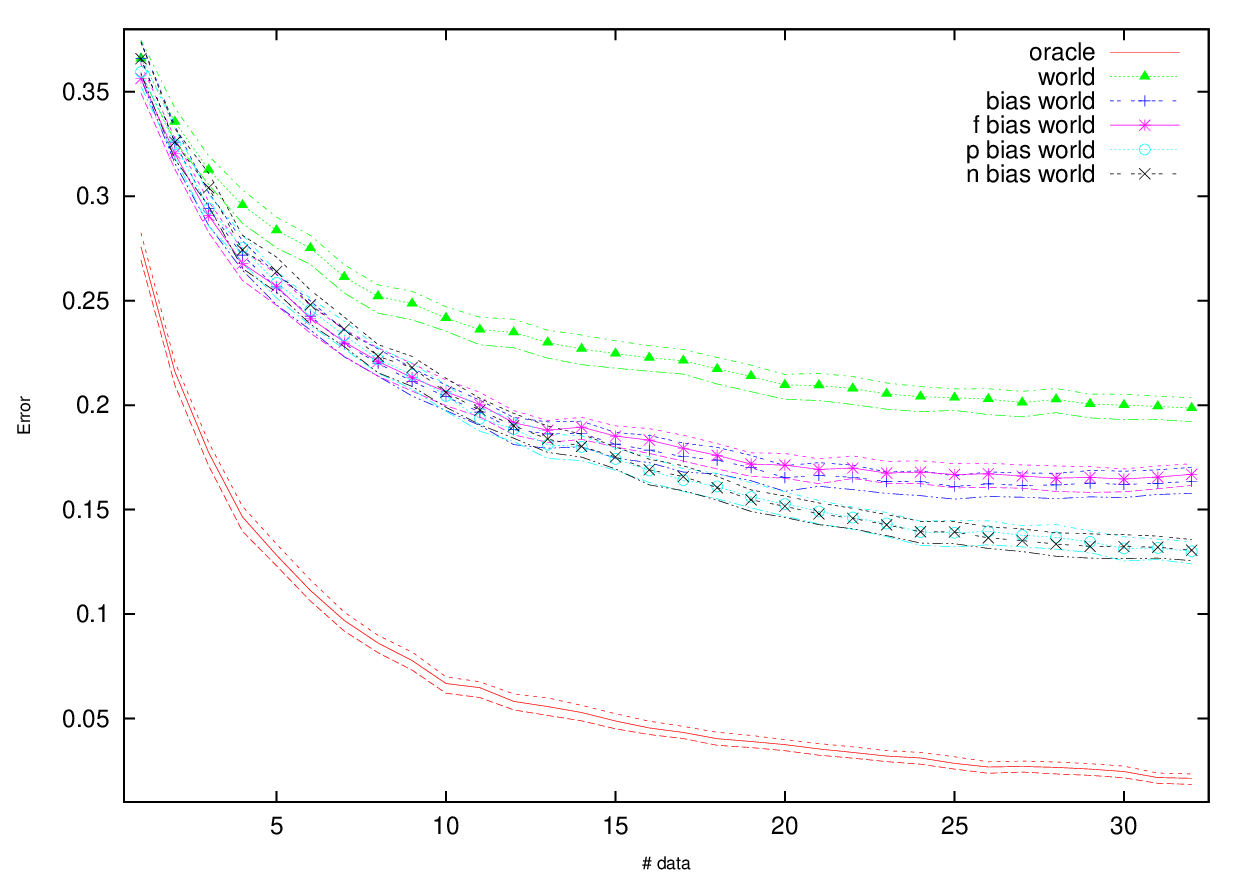}
  \caption{The evolution of error rates as more data becomes
    available, when the user model and prior are estimated.  The
    points indicate means from $10^4$ runs and the lines top and
    bottom $5\%$ percentiles from a bootstrap sample.}
  \label{fig:synthetic_user_bootstrap}
\end{figure}

\section{Experimental evaluation}
\label{sec:experiments}
We have performed a number of experiments in order to evaluate the
proposed approach and compared it to the full Bayesian version of the
well-known world model approach.  We performed a set of experiments on
synthetic data, and another set of experiments on real data.


For the synthetic experiments, we assume multinomial models, but
rather than knowing $\gamma$, we use data from other users to form an
empirical estimate $\hat{\gamma}$, as described in
Sec.~\ref{sec:multinomial-estimation}. Furthermore, $q$ is itself unknown and is
estimated via Bayesian updating from $\hat{\gamma}$ and some data
specific to the user.  We then compare the oracle and the world model
approach (based on $\gamma$) with a number of differently biased
adversary models. The world model is based on the estimate
$\hat{\gamma}$. The adversary model uses the world model
$\hat{\gamma}$ as the adversary prior ($\xi$).
 
The second group concerns experiments on data gathered from an access
control system.  The data has been discretized into $1320$ integer
variables, in order for it to be modelled with multinomials. The
models are of course not available so we must estimate the priors:
The data of a subset of users is used to estimate $\hat{\gamma}$. The
remaining users alternatively take on the roles of legitimate users
and adversaries.

We compare the following types of models, which correspond to the
legends in the figures of the experimental results.
\begin{inparaenum}[(a)]
\item The {\bf oracle} model, which enjoys perfect information
  concerning adversary and user distributions.
\item The {\bf world} model, which uses the prior over user models as
  a surrogate for the adversary model.
\item The {\bf bias world} model, which uses all but the last observation to obtain a posterior over adversary models, and similarly:
\item the {\bf f bias world} model, which uses all observations,
\item the {\bf p bias world} model, which weighs the observations by $1/2$ and
\item the {\bf n bias world} model, which uses the first half of the observations.
\end{inparaenum}
In all cases, we used percentile calculations based on multiple runs
and/or bootstrap replicates~\cite{Efron:1993:Introduction} to assess
the significance of results.

\subsection{Synthetic experiments}

For this evaluation, we ran $10^4$ independent experiments and
employed multinomial models. For each experiment, we first generated
the true prior distribution over user models $\gamma$. This was
created by drawing Dirichlet parameters $\phi_i$ independently from a
Gamma distribution.  We also generated the true prior distribution
over adversary models $\gamma'$, by drawing from the same Gamma
distribution.  Then, a user model $q$ was drawn from $\gamma$ and an
adversary model $w$ was drawn from $\gamma'$. Finally, by flipping a
coin, we generated data $x_1, \ldots, x_n$ from either $q$ or
$w$. Assuming equal prior probabilities of user and adversary, we
predicted the most probable class and recorded the error. This was
done for all subsequences of the observations' sequence $x$. Thus, the
experiment measures the performance of methods when the amount of data
that informs our decision increases.

For these experiments, we estimate the actual Dirichlet distribution
with $\hat{\gamma}$.  This estimation is performed via empirical Bayes
using data from $1000$ users drawn from the actual prior $\gamma$.  At
the $k$-th run, we draw a user model $q_k \sim \gamma$ and
subsequently draw $x_k \sim q_k$. We then use $\hat{\gamma}$ and the
user data $x_k \in \Naturals^K$, to estimate a posterior over user
models for the $k$-th user, $\psi_k(q) \defn \hat{\gamma}(q|x_k)$. The
estimated prior $\hat{\gamma}$ is also used as the world model and as
the prior over adversary models.  The results, shown in
Figure~\ref{fig:synthetic_user_bootstrap}, show that the biased models
consistently outperform the classic world model approach, while the
partially biased models become significantly better than the fully
biased models when the amount of observations increases. This is
encouraging for application to real-world data.

\subsection{Real data}
The real world data were collected from an RFID based access control
system used in two buildings of the TNO organization (Netherlands
Organization for Applied Scientific Research).  The data were
collected during a three and a half month period, and they include
successful accesses of 882 users, collected from 55 RFID readers
granting access to users attempting to pass through doors in the
buildings.

The initial data included three fields: the time and date that the
access has been granted, the reader that has been used to get access
and the ID of the RFID tag used\footnote{The data were sanitised to
  avoid privacy issues.}.  In order to use the data in the
experimental evaluation of the proposed model framework, we have
discretized the time into hour-long intervals, and counted the number
of accesses, per hour, per door for each user, in each day. This
resulted in a total of $\approx 2\cdot10^5$ records. Since there are
24 hour-long slots in a day, and a total of 55 reader-equipped doors,
this discretisation allowed us to model each user by a 1320-degree
multinomial/Dirichlet model. Thus, even though the underlying
Dirchlet/multinomial model framework is simple, the very high
dimensionality of the observations makes the estimation and decision
problem particularly taxing.

\subsubsection{Experiments}
\begin{figure}[htb]
  \centering
  \includegraphics[width=\textwidth]{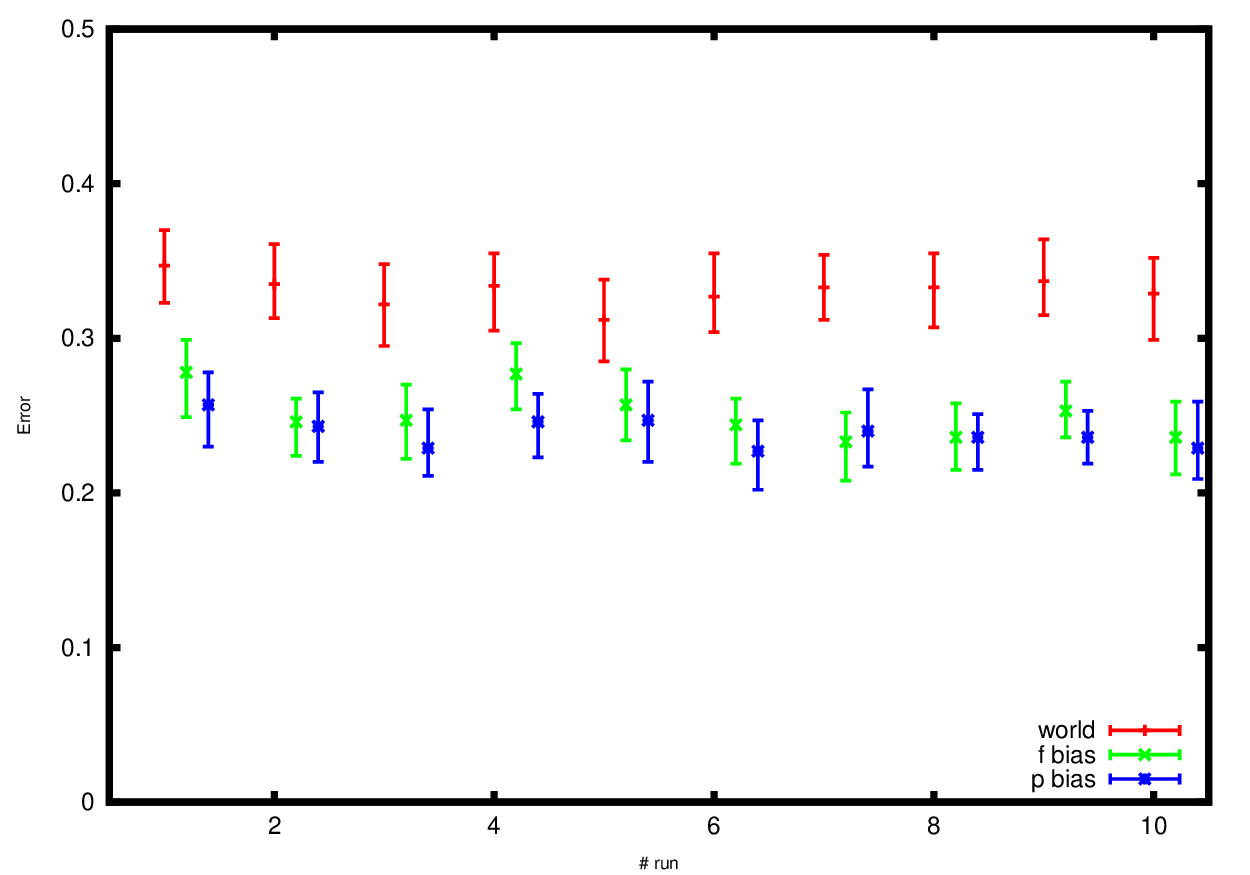}
  \caption{Error rates for 10 runs on the TNO door data. The error
    bars indicate top and bottom $5\%$ percentiles from $100$ bootstrap
    samples from $10^3$ repetitions per run.}
  \label{fig:door_bootstrap}
\end{figure}

We performed 10 independent runs. For the $k$-th run, we selected a
random subset $U_\gamma$ of the complete set of users $U$, such that
$|U_\gamma|/|U| = 2/3$.  We used $U_\gamma$ to estimate the world
model $\hat{\gamma}$. The remaining users $U_T = U \backslash
U_\gamma$ were used to estimated the error rate over $10^3$
repetitions. For the $j$-th repetition, we randomly selected a user $i
\in U_T$ with at least $10$ records $D_i$. We used half of those
records, $\bar{D}_i$, to obtain $\psi_i(q) \defn \hat{\gamma}(q|\bar{D}_i)$. By
flipping a coin, we obtain either
\begin{inparaenum}[(a)]
  \item one record from $D_i \backslash \bar{D}_i$, or 
  \item data from
    some other user in $U_T$.
\end{inparaenum}
Let us call that data $x_j$. For the biased models, we set $\xi =
\hat{\gamma}$ and then used $x_j$ to obtain $\xi(w|f(x_j))$, where
$f(\cdot)$ denotes the appropriate
transformation. Figure~\ref{fig:door_bootstrap} shows results for the
baseline world model approach ({\bf world}), where $f(x) =\emptyset$,
as the unmodified world model is used for the adversary, the full bias
approach ({\bf f bias}), where $f(x) = x$ since all the data is used,
and finally the partial bias approach ({\bf p bias}) where $f(x) =
x/2$. The other approaches are not examined, as the oracle is not
realisable, while the half-data and the all-but-last-data biased
models are equivalent to the baseline world model, since we do not
have a sequence of observations, but only a single record.

As can be seen in Figure~\ref{fig:door_bootstrap}, the baseline world
model is always performing worse than the biased models, though in two
runs the full bias model is close.  Finally, though the two biased
models are not distinguishable performance-wise, we noted a difference
in the ratio of false positives to false negatives. Over the 10 runs,
this was $0.2 \pm 0.1$ for the world model approach, $2.5 \pm 0.5$
for the fully biased model, and $0.9 \pm 0.2$ for the partially biased
model.

\section{Conclusion}

We have presented a very simple, yet effective approach for
classification problems where one class has no data.  In particular,
we define a prior over models which can be estimated from population
data. This is adapted, as in the standard world-model approach, to a
specific user.  We introduce the idea of creating an adversary model,
for which no {\em labelled} data exists, from the prior and currently
seen data. Within the subjective Bayesian framework, this allows us to
obtain a subjective upper bound on the probability of an attack.

Experimentally, it is shown~\footnote{In an accompanying technical
  report~\cite{dimitrakakis:tr-uva-09-02}, the effect of
  dimensionality on the performance of the method is also
  examined. There, it is shown that a Bayesian framework is essential
  for such a scheme to work and that naive approaches perform
  progressively worse as the dimensionality increases. }
that:
\begin{inparaenum}[(a)]
\item we outperform the classical world model approach, while
\item it is always better to only {\em partially} condition the models on the new observations.
\end{inparaenum}

It is possible to extend the approach to the cost-sensitive case.
Since we already have bounds on the probability of each class,
together with a given cost matrix, we can also calculate bounds on the
expected cost. This will allow us to make cost-sensitive decisions.

A related issue is whether to alter the {\em a priori class
  probabilities}; in our comparative experiments we used equal fixed
values of 0.5. It is possible to utilise the population data to tune
it in order to achieve some desired false positive / negative
ratio. Such an automatic procedure would be useful for an expected
performance curve~\cite{Bengio:Expected:2005} comparison between the
various approaches. Finally, since the experiments on this relatively
complex problem gave promising results, we plan to evaluate it on
other problems that exhibit a lack of adversarial data.

\iftrue
 \appendix
 \section{Proof}

\label{app:proofs}
\begin{proof}[Lemma~\ref{lem:posterior-inequality}]
  For discrete $\CM$, the marginal prior $\xi(x)$ can be re-written as
  follows:
  \begin{equation}
    \xi(x) = \sum_\mu \xi(x, \mu)
    = \sum_\mu \xi(x|\mu) \xi(\mu)
    = \sum_\mu \mu(x) \xi(\mu),
  \end{equation}
  and similarly: $ \xi'(x) = \frac{1}{\sum_\mu \mu(x) \xi(\mu)}
  \sum_\mu \mu(x)^2 \xi(\mu)$.
  Thus, to prove the required statement, it is sufficient to show
  \begin{align}
    \left(\sum_\mu \mu(x)^2 \xi(\mu)\right)^{1/2} \geq \sum_\mu \mu(x)\xi(\mu).
  \end{align}
  Similarly, for continuous $\CM$, we obtain:
  \begin{align}
    \left(\int \mu(x)^2 \, d\xi(\mu)\right)^{1/2} \geq \int \mu(x) \, d\xi(\mu).
  \end{align}
  In both cases, the norm induced by the probability measure $\xi$ on
  $\CM$ is $\|f\|_2 = (\int_\CM |f(\mu)|^2 \, d\xi(\mu))^{1/2}$, thus allowing
  us to included apply the Cauchy-–Schwarz inequality $\|fg\|_1 \leq
  \|f\|_2 \|g\|_2$. By setting $f(\mu) = \mu(x)$ and $g(\mu) = 1$, we
  obtain the required result, since $\|g\|_2 = (\int_\CM  d\xi(\mu))^{1/2} = 1$, as
  $\xi$ is a probability measure.  
\end{proof}

\section*{Acknowledgements}
This work was supported by the Netherlands Organization for Scientific
Research (NWO) under the RUBICON grant "Intrusion Detection in
Ubiquitous Computing Technologies" and the ICIS project, supported by
the Dutch Ministry of Economic Affairs, grant nr: BSIK03024.

\bibliographystyle{plainnat}
\bibliography{attacker}

\end{document}